\documentclass[10pt,twocolumn,letterpaper]{article}

\usepackage{iccv}
\usepackage{graphicx}
\usepackage{times}
\usepackage{epsfig}
\usepackage{amsmath}
\usepackage{amssymb}
\usepackage{algorithm}
\usepackage{multirow}
\usepackage{mathtools,xparse}
\usepackage{siunitx}
\usepackage{stmaryrd}
\usepackage{algorithmicx}
\usepackage[noend]{algpseudocode}
\usepackage{placeins}
\usepackage{mdframed}
\usepackage{amsmath,amssymb} 
\usepackage{color}
\usepackage{subcaption}
\usepackage[table]{xcolor}

\DeclarePairedDelimiter{\ceil}{\lceil}{\rceil}
\newcommand{\Point}{\textbf{p}}
\newcommand{\trans}{\text{T}}
\newcommand{\ph}{\phantom{x}}
\newcommand{\usac}{USAC$^*$ }

\definecolor{g1}{rgb}{1.00,1.00,1.00}

\usepackage[pagebackref=true,breaklinks=true,letterpaper=true,colorlinks,bookmarks=false]{hyperref}

\iccvfinalcopy 

\def\iccvPaperID{5019} 

\ificcvfinal\fi
\begin{document}

\title{Progressive NAPSAC: sampling from gradually growing neighborhoods}

\author{Daniel Barath$^{12}$, Maksym Ivashechkin$^{1}$, and Jiri Matas$^{1}$\\
$^1$ Centre for Machine Perception, Department of Cybernetics \\
  Czech Technical University, Prague, Czech Republic \\
  $^2$ Machine Perception Research Laboratory, 
  MTA SZTAKI, Budapest, Hungary \\
    {\tt\small barath.daniel@sztaki.mta.hu}
}


\maketitle

\begin{abstract}
We propose Progressive NAPSAC, P-NAPSAC in short,
which merges the advantages of local and global sampling by drawing samples from gradually growing neighborhoods.
Exploiting the fact that nearby points are more likely to originate from the same geometric model, P-NAPSAC finds local structures earlier than global samplers.
We show that the progressive spatial sampling in P-NAPSAC can be integrated with  PROSAC sampling, which is applied to the first, location-defining, point. 
P-NAPSAC is embedded  in USAC~\cite{raguram2013usac}, 
a state-of-the-art robust estimation pipeline, which we
further improve by implementing its local optimization as in Graph-Cut RANSAC~\cite{barath2018graph}. We call the resulting estimator \usac. 

The method is tested on homography and fundamental matrix fitting on a total of $10\;691$ models from seven publicly available datasets. \usac with P-NAPSAC outperforms reference methods in terms of speed on all problems.
\end{abstract}

\section{Introduction}

The RANSAC (RANdom SAmple Consensus) algorithm proposed by Fischler and Bolles~\cite{fischler1981random} has become the most widely used robust estimator in computer vision. 
RANSAC and its variants have been successfully applied to a wide range of vision tasks, e.g., motion segmentation~\cite{torr1993outlier}, short baseline stereo~\cite{torr1993outlier,torr1998robust}, wide baseline stereo matching~\cite{pritchett1998wide,matas2004robust,mishkin2015mods}, detection of geometric primitives~\cite{sminchisescu2005incremental}, image mosaicing~\cite{ghosh2016survey}, and to perform~\cite{zuliani2005multiransac} or initialize multi-model fitting~\cite{isack2012energy,pham2014interacting}.
In brief, vanilla RANSAC repeatedly selects minimal random subsets of the input point set and fits a model, e.g., a line to two 2D points or a fundamental matrix to seven 2D point correspondences.
Next, the quality of the estimated model is measured, for instance by the cardinality of its support, i.e., the number of inlier data points. Finally, the model with the highest quality, polished, e.g. by least squares fitting  of all inliers, is returned.

\begin{figure}[h]
    \centering
    \includegraphics[width=1.0\columnwidth]{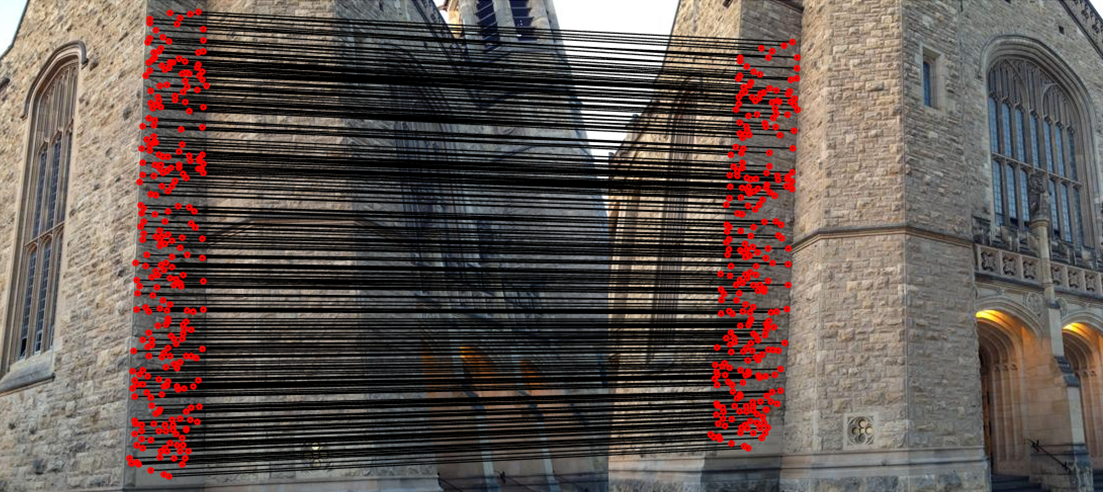}
    \caption{The Bonhall image pair from the {\fontfamily{cmtt}\selectfont AdelaideRMF homography} dataset.
    Given $866$ correspondences as input, 
    PROSAC~\cite{chum2005matching} found $58$ and Progressive NAPSAC all the $62$ inliers. 
    To estimate the homography, PROSAC tested $99\;001$ four-tuples of correspondences in $1.26$ seconds while P-NAPSAC tested $11\;605$ four-tuples in $0.15$ seconds (on average, over $100$ runs). Inlier correspondences are marked by a line segment joining the corresponding points.}
    \label{fig:example_result}
\end{figure}

Since the publication of RANSAC, many modifications have been proposed, improving all components of the algorithm. 
For instance, MLESAC~\cite{torr2000mlesac} estimates the model quality by a maximum likelihood process with all its beneficial properties, albeit under certain assumptions about inlier and outlier distributions. 
In practice, MLESAC results are often superior to the inlier counting of plain RANSAC, and they are less sensitive to the user-defined inlier-outlier threshold.
In MSAC~\cite{torr2002bayesian}, the robust estimation is formulated as a process that estimates both the parameters of the data distribution and the quality of the model in terms of maximum a posteriori. 

\begin{figure}[h]
    \centering
	\begin{subfigure}[t]{0.495\columnwidth}
	    \includegraphics[width=1.0\columnwidth]{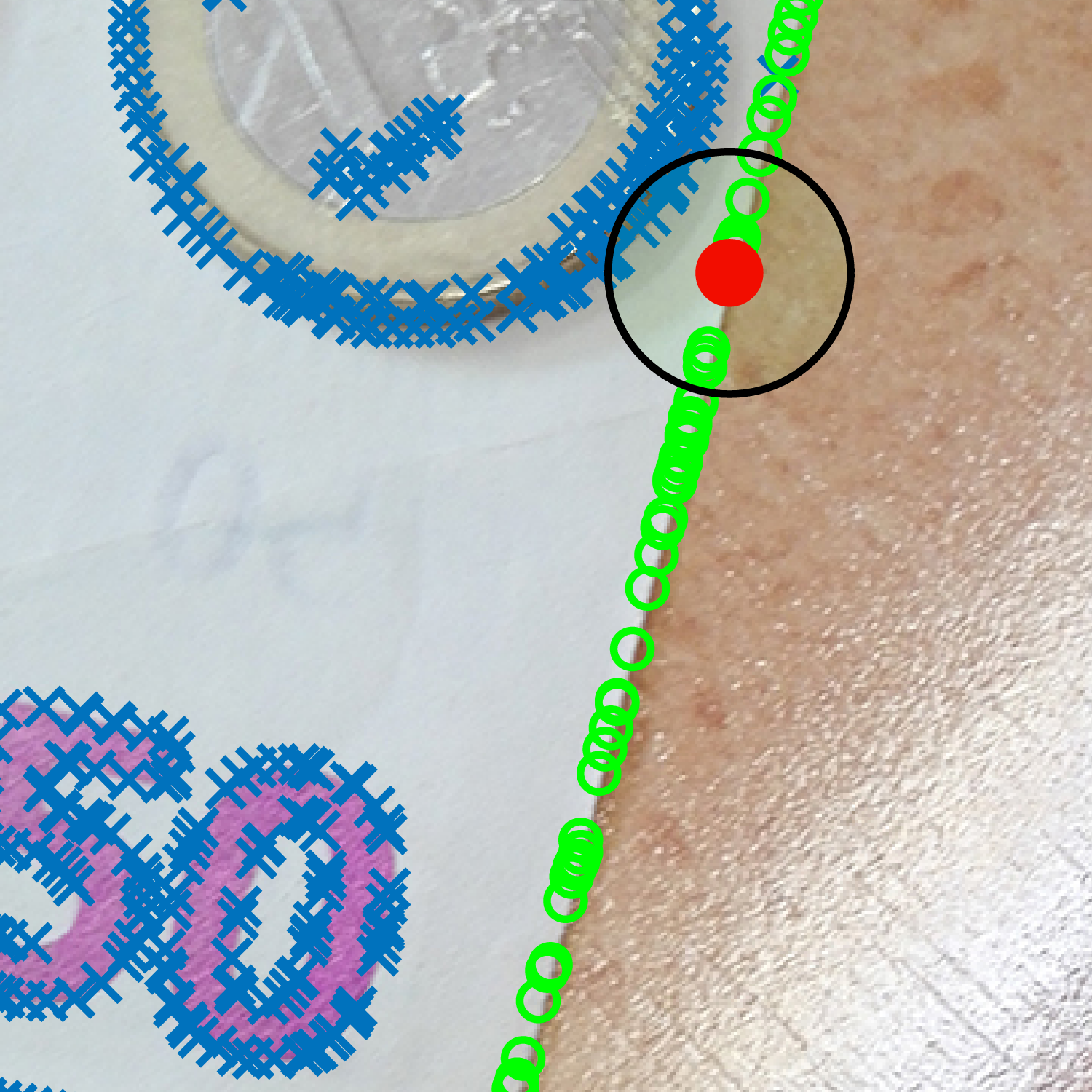}
	    \caption{}
	\end{subfigure}\hfill
	\begin{subfigure}[t]{0.495\columnwidth}
	    \includegraphics[width=1.0\columnwidth]{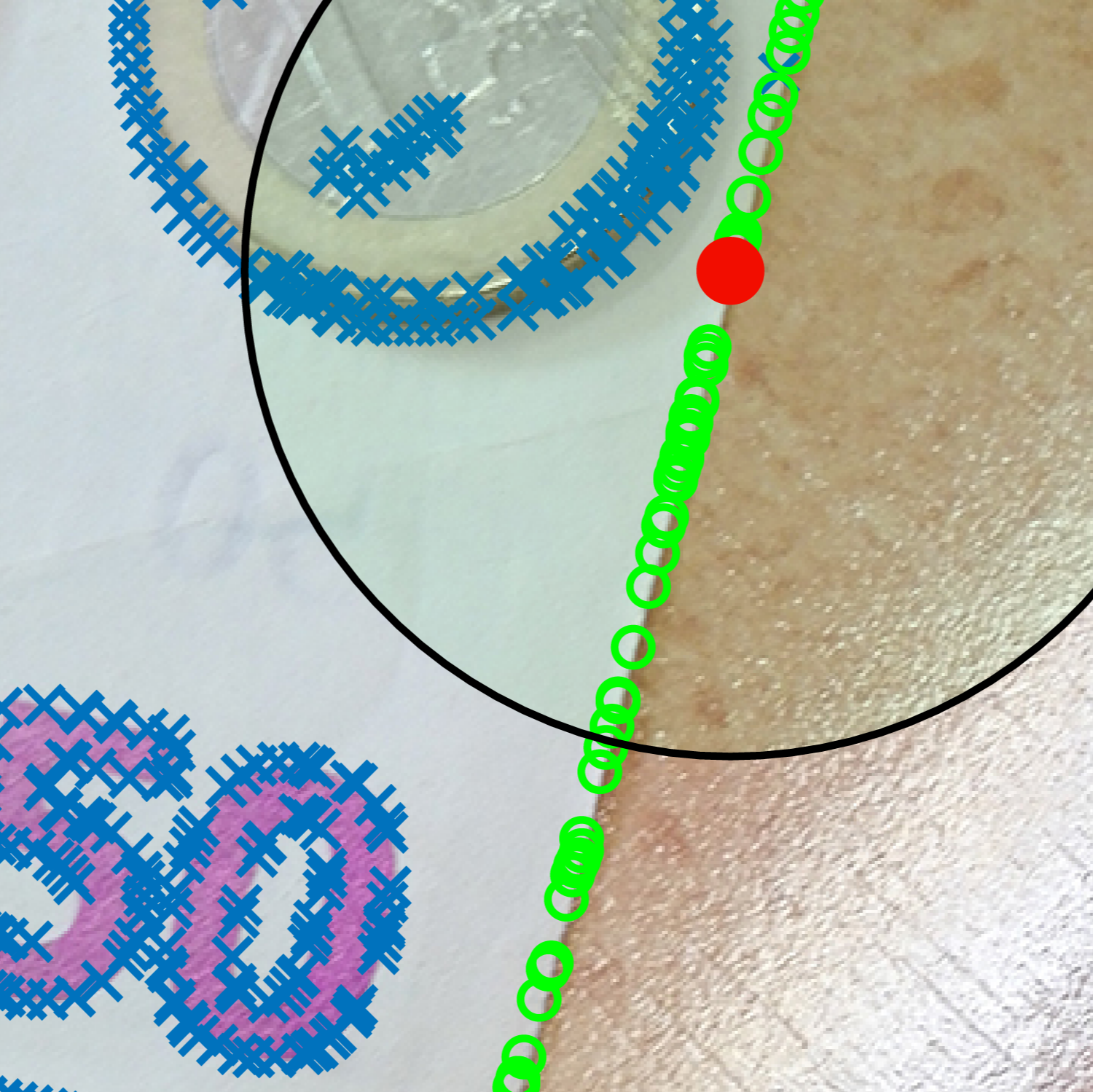}
	    \caption{ }
	\end{subfigure}
        \caption{\textit{Local inlier ratio.} 
        The green circles are the inliers of the sought line and the blue crosses are outliers.
        The red dot is $\Point_i$, an inlier selected randomly, or using the PROSAC sampler. 
        The circle around $\Point_i$  determines the neighborhood of $\Point_i$. 
        (a) Radius $r = 50$ px. Within the circle, the inlier number is $|\mathcal{I}| = 11$, point number is $|\mathcal{S}_{i,r}| = 20$ and  inlier ratio is $|\mathcal{I}| / |\mathcal{S}_{i,r}| = 0.55$.
        (b) Radius $r = 150$ px. Within the circle, $|\mathcal{I}| = 48$, $|\mathcal{S}_{i,r}| = 166$ and the inlier ratio is $|\mathcal{I}| / |\mathcal{S}_{i,r}| = 0.29$. }
    \label{fig:napsac_radius_sampling}
\end{figure}

Observing that RANSAC requires in practice  more samples than theory predicts, Chum et al.~\cite{chum2003locally,lebeda2012fixing} identified a problem that not all all-inlier samples are ``good'', i.e., lead to a model accurate enough to distinguish all inliers, e.g., due to poor conditioning of the selected random all-inlier sample. They addressed the problem by introducing the locally optimized RANSAC (LO-RANSAC) that augments the original approach with a local optimization step applied to the \textit{so-far-the-best} model. 
This approach had been improved in Graph-Cut RANSAC~\cite{barath2018graph} which takes into account the fact that real-world data often form spatially coherent structures.
Therefore, it considers the proximity of the points in the local optimization which leads to superior results.

Methods for reducing the dependency on the inlier-outlier threshold include
MINPRAN~\cite{stewart1995minpran} which assumes that the outliers are uniformly distributed and finds the model where the inliers are least likely to have occurred randomly. 
Moisan et al.~\cite{moisan2012automatic} proposed a contrario RANSAC, to optimize each model by selecting the most likely noise scale.
Barath et al.~\cite{barath2019magsac} proposed a method marginalizing over the possible noise scales in order to eliminate the threshold from the model quality calculation.

NAPSAC~\cite{nasuto2002napsac} and PROSAC~\cite{chum2005matching} modify the standard RANSAC sampling strategy of selecting points at random, to increase the probability of selecting an all-inlier sample early. 
PROSAC exploits an a priori predicted inlier probability rank of the points and starts the sampling with the most promising ones.
Progressively, samples which are less likely to lead to the sought model are drawn. 
PROSAC and other RANSAC-like samplers treat models without considering
that inlier points often in the proximity of each other. 
This approach is effective when finding a global model with inliers sparsely distributed in the scene, for instance, the rigid motion induced by changing the viewpoint in two-view matching.
However, as it is often the case in real-world data, if the model is localized with inlier points close to each other, robust estimation can be significantly speeded by exploiting this in sampling.

NAPSAC assumes that inliers are spatially coherent.
It draws samples from a hyper-sphere centered at the
first, randomly selected, point. 
If this point is an inlier, the rest of the points sampled in its proximity are more likely to be inliers than the points outside the ball.
The localized sampling of NAPSAC leads to fast, successful termination in many cases. 
However, it suffers from a number of issues in practice.
First, the models fit to local all-inlier samples are often too imprecise for distinguishing all inliers in the data due to the bad conditioning of the points.
Second, in some cases, estimating a model from a localized sample leads unavoidably to degenerate solutions.
For instance, when fitting a fundamental matrix by the seven-point algorithm, the set of correspondences must originate from more than one plane.
Therefore, there is a trade-off between near, likely all-inlier, and global, well-conditioned, lower all-inlier probability samples.
Third, when the points are sparsely distributed and not spatially coherent, NAPSAC often fails to find the sought model. 

In this paper, we propose Progressive NAPSAC (P-NAPSAC) which merges the advantages of local and global sampling by drawing samples from gradually growing neighborhoods.
Considering that nearby points are more likely to originate from the same geometric model, P-NAPSAC finds local structures earlier than global samplers.
In addition, it does not suffer from the weaknesses of purely localized samplers due to progressively blending from local to global sampling, where the blending factor is a function of the input data. 
Moreover, P-NAPSAC is included in a state-of-the-art robust estimation pipeline, i.e., USAC~\cite{raguram2013usac}, including PROSAC sampling applied to the first, location-defining, point. It is combined with Graph-Cut RANSAC~\cite{barath2018graph} leading to \usac. 
The method was tested on homography and fundamental matrix fitting on a total of $10\;691$ models from seven publicly available datasets -- P-NAPSAC leads to faster termination of \usac than state-of-the-art samplers. 

An example homography estimation problem is shown in Figure~\ref{fig:example_result}.
PROSAC found $58$ and Progressive NAPSAC all the $62$ inliers, while being about eight times faster.

\section{N Adjacent Points SAmple Consensus}

In this section, we briefly discuss the N Adjacent Points SAmple Consensus (NAPSAC) sampling technique~\cite{nasuto2002napsac}.
NAPSAC builds on the assumption that the points of a model are spatially structured and, thus, sampling from local neighborhoods increases the inlier ratio locally. In brief, the algorithm is as follows:
\begin{enumerate}
    \item Select an initial point $\Point_i$ randomly from all points. 
    \item Find the set $\mathcal{S}_{i, r}$ of points lying within a hyper-sphere of radius $r$ centered on $\Point_i$.
    \item If the number of points in $\mathcal{S}_{i, r}$ is less than the minimal sample size then restart from step 1. 
    \item Select points from $\mathcal{S}_{i, r}$ uniformly until the minimal set has been selected, inclusive of $p_i$.
\end{enumerate}
Note that when using the k-nearest-neighbors algorithm to determine the neighborhood structure, $r$ is replaced by $k$ and we denote the implied neighborhood by $\mathcal{S}_{i, k}$.

This results in a cluster of points being selected from a ball. 
If the initial point $\Point_i$ lies on the model, then the rest of the points sampled adjacently are theoretically more likely to be inliers than the points outside the ball.
If there are not enough points within the hyper-sphere to estimate the model, then the sample is considered a failure. 
 
\begin{table}
	\center
	\resizebox{0.999\columnwidth}{!}{\begin{tabular}{ c  l | c  l }
    \hline
   		\multicolumn{4}{ c }{ \cellcolor{black!10}Notation\rule{0pt}{2.0ex} } \\[0.2mm]
    \hline 
   		$\mathcal{P}$ & - Set of data points & 
   		$\Point_i$ & - $i$-th point in $\mathcal{P}$ \\ 
   		$\theta$ & - Model parameters &
   		$m$ & - Minimal sample size \\
   		$\mathcal{I}$ & - Set of inliers &
   		$\mathcal{S}_{i,k}$ & - $k$ closest neighbors of $\Point_i$ \; \\
   		$\mathcal{M}_{i,j}$ & - $j$-th sample containing $\Point_i$ & 
   		$x_{i,j}$ & - Indices ($\in [1, |\mathcal{P}|]^m$) in $\mathcal{M}_{i,j}$ \\
    \hline     
\end{tabular}}
\label{tab:notation}
\end{table}

Fig.~\ref{fig:napsac_radius_sampling} demonstrates the advantage of sampling locally.
The image shows banknotes and coins. 
The green circles are inliers of the sought line and the blue crosses are outliers.
The red dot is $\Point_i$, an inlier selected randomly. 
The circle around $\Point_i$ is the one used for determining the neighborhood of $\Point_i$. 
It can be seen that bigger the circle (i.e., the size of neighborhood), lower the inlier ratio. 

There nevertheless are three major issues of local sampling in practice. 
\textit{First}, it was observed that models fit to local all-inlier samples are often too imprecise for distinguishing all inliers in the data.
This is caused by the bad conditioning of the noisy inliers. 
Addressing this problem, Chum et al.~\cite{chum2003locally} proposed LO-RANSAC applying an iterated least-squares fitting to the inliers of the current model.
\textit{Second}, in some cases, estimating a model from a localized sample leads unavoidably to degeneracy. 
For instance, when fitting a fundamental matrix by the seven-point algorithm, the set of correspondences must originate from more than one plane.
This usually means that the correspondences are beneficial to be far from each other. 
Therefore, purely localized sampling fails. 
\textit{Third}, in the case of having global structures, e.g., the rigid motion of the background in an image sequence, local sampling postpones the termination further than what it would be by considering all points in the sampling at once and not just local neighborhoods.
We, therefore, propose a transition between local and global sampling progressively blending from one into the other.

\section{Progressive NAPSAC}

In this section, a method called Progressive NAPSAC is proposed combining the strands of NAPSAC-like local sampling and the global sampling of RANSAC. 
The idea of Progressive NAPSAC is as follows: first, sample uniformly from a local subset of the data points. 
Then add new points gradually to the subset and, thus, blend into the global sampling of RANSAC. 
The subset of data points which are used for drawing the samples, unlike in PROSAC~\cite{chum2005matching}, are not selected from the data based on the quality, but by considering the neighborhoods of the data points independently. 
A hyper-sphere is assigned to each point and its radius is gradually increased.
In the case of localized models, the samples are more likely to contain inliers solely and, therefore, trigger early termination. 
When the points of the sought model do not form spatially coherent structures, the gradual increment of the radii of the neighborhood balls leads to finding global structures not noticeably later than by using global samplers, e.g., PROSAC. 

\subsection{The growth function and sampling}

The design of the growth function that defines how fast the investigated neighborhood grows around a selected point $\Point_i$ must find the balance between the strict NAPSAC assumption, i.e., the models are entirely localized, and the RANSAC approach that treats every model on a global scale. 
The problem is similar to that of PROSAC where the algorithm is balancing between using and not using an a priori determined heuristics which measures the quality of a particular sample. 
In the current case, where the models are assumed to be localized, the distances from a point can be considered as quality measures in the PROSAC scoring. 

Let $ \{ \mathcal{M}_{i,j} \}_{j = 1}^{T_n} = \{ \Point_{i}, \Point_{x_{i,j,1}}, ..., \Point_{x_{i, j, m - 1}} \}_{j = 1}^{T_n}$ denote the sequence of samples $\mathcal{M}_{i,j} \subset \mathcal{P}^*$ starting with point $\Point_i \in \mathcal{P}$ and drawn uniformly by RANSAC, where $m$ is the sample size, $T_n$ is the number of all RANSAC samples, $\mathcal{P}^*$ is the power set of $\mathcal{P}$, and $x_{i, j, 1}$, $...$, $x_{i, j, m - 1} \in \mathbb{N}^+$ are indices, each referring to a particular point in the point set. 
Let $ \{ \mathcal{M}_{(i,j)} \}_{j = 1}^{T_n}$ be a sequence of the same samples sorted in ascending order according to the sum of distances of the contained points from the $i$-th one as follows:
\begin{equation*}
    j < k \Rightarrow \sum_{s = 1}^{m - 1} |\Point_{x_{i,j,s}} - \Point_i| \leq \sum_{s = 1}^{m - 1} |\Point_{x_{i,k,s}} - \Point_i|,
\end{equation*}
where $|.|$ is a norm, e.g., the Euclidean distance for 2D points, measuring the distance of two points.
If the samples are taken in order $\mathcal{M}_{(i,j)}$, the samples which consist of points close to the $i$-th one are drawn early.
Progressively, samples which contain data points farther from the $\Point_i$ are drawn. 
After $T_n$ samples, exactly all RANSAC samples $\{ \mathcal{M}_{i,j} \}_{j = 1}^{T_n}$ are drawn. 

Since the problem is quite similar to that of PROSAC, the same growth function can be used after considering that the first point $\Point_i$ in the sample is selected already. 
Let $T_k$ be an average number of samples from $\{ \mathcal{M}_{i,j} \}_{j = 1}^{T_n}$ that contain $\Point_i$ and the other points are from $\mathcal{S}_{i,k}$ only as follows:
\begin{equation*}
    T_k  = T_n \frac{\binom{k}{m - 1}}{\binom{n}{m - 1}}  = T_n \prod_{i = 0}^{m - 2} \frac{k - i}{n - i},
\end{equation*}
where $n$ is the number of data points. After rearranging the equation and calculating the $T_{k + 1} / T_k$ ratio, $T_{k + 1}$ can be recursively defined as
\begin{equation*}
    T_{k + 1} = \frac{k + 1}{k + 2 - m} T_k.
\end{equation*}
Since the values are not integers, we define $T_k' = 1$ and $T_{k + 1}' = T_k' + \ceil{T_{k + 1} - T_k}$. 
The growth function is then defined as follows:
\begin{equation*}
    q(t_i) = \min \{ k \; : \; T_k' \geq t_i \},
\end{equation*}
where $t_i$ is the number of drawn samples which include the $i$-th point. 
The $t_i$-th sample $\mathcal{M}_{i, t_i}$ containing $\Point_i$ consists of
\begin{equation*}
    \mathcal{M}_{i,t_i} = \{ \Point_{i}, \Point_{i,g(t_i)} \} \cup \mathcal{M}_{i,t_i}',
\end{equation*}
where $\mathcal{M}_{i,t_i}' \subset \mathcal{S}_{i, g(t_i) - 1}$ is a set of $|\mathcal{M}_{i,t_i}'| = m - 2$ data points, excluding $\Point_i$, randomly drawn from $\mathcal{S}_{i, g(t_i) - 1}$. $\Point_{i, g(t_i)}$ is the $g(t_i)$-th point if the points are ordered with respect to their distances from $\Point_i$.

\textit{Growth of the iteration number.} Given point $\Point_i$, the corresponding $t_i$ is increased in two cases. 
First, $t_i \leftarrow t_i + 1$ when $\Point_i$ is selected to be the center of the hyper-sphere.
Second, $t_i$ is increased when $\Point_l$ is selected, the neighborhood of $\Point_l$ contains $\Point_i$ and, also, that of $\Point_i$ contains $\Point_l$. 
Formally, let $\Point_l$ be selected as the center of the sphere ($l \neq i \; \wedge \; l \in [1, n]$). 
Let sample $\mathcal{M}_{l,j} = \{ \Point_l, \Point_{x_{l, j, 1}}, ..., \Point_{x_{l, j, m - 1}}  \}$ be selected randomly as the sample in the previously described way. If $i \in \{ x_{l, j, 1}, ..., x_{l, j, m - 1} \}$ (or equivalently, $\Point_i \in \mathcal{M}_{l,j}$) and $\Point_l \in \mathcal{S}_{i,g(t_i)}$ then $t_i$ is increased by one. 

The algorithm (shown in Alg.~\ref{alg:post_processing}) can be imagined as a PROSAC sampling defined for every $i$-th point independently, where the sequence of samples for the $i$-th point depends on its neighbors. 
After the initialization, the first main step of Alg.~\ref{alg:post_processing} is to select $\Point_i$ as the center of the sphere and update the corresponding $t_i$.
Then a semi-random sample is drawn consisting of the selected $\Point_i$, $m - 2$ random points from $\mathcal{S}_{i, k_i - 1}$ (i.e., the points in the sphere around $\Point_i$ excluding the farthest one) and $\Point_{i, k_i}$ which is the farthest point in the sphere.
Based on the random sample, the corresponding $t$ values are updated. 
Finally, the implied model is estimated, and its quality is measured.

\begin{algorithm}
\begin{algorithmic}[1]
	\Statex{\hspace{-1.0em}\textbf{Input:} $\mathcal{P}$ -- points; $\mathcal{S}$ -- neighborhood structure}
    \Statex{$t_1, ..., t_n := 0$, $k_1, ..., k_n := m$}
   	\Statex{Repeat the followings until the solution is found.}
    \Statex{\hrulefill}
   	\Statex{\textbf{Selection of the sphere center}:}
   	\State{Let $\Point_i$ be a random point.}\Comment{e.g., selected by PROSAC.}
   	\State{$t_i := t_i + 1$}
   	\If{($t_i = T_{k_i}' \wedge k_i < n$)}
   	    \State{$k_i := k_i + 1$}
   	\EndIf
    \Statex{\hrulefill}
   	\Statex{\textbf{Semi-random sample $\mathcal{M}_{t_i}$ of size $m$}:}
   	\If{$T_{k_i}' < t_i$}
       	\State{The sample contains $\Point_i$, $m - 2$ points selected from}
        \Statex{\hskip\algorithmicindent $\mathcal{S}_{i, k_i - 1}$ at random and $\Point_{i, k_i}$.}
   	\Else
   	    \State{Select $m$ points from $\mathcal{P}$ at random.}
   	\EndIf
    \Statex{\hrulefill}
   	\Statex{\textbf{Increase the iteration number}:}
   	\For{$\Point_j \in \mathcal{M}_{t_i}$}
   	    \If {$\Point_i \in \mathcal{S}_{j, k_j}$}
   	        \State{$t_j := t_j + 1$}
   	    \EndIf
   	\EndFor
   	\Statex{\textbf{Model parameter estimation}}
   	\State{Compute model parameters $\theta$ from sample $\mathcal{M}_{t_i}$.}
   	\Statex{\textbf{Model verification}}
   	\State{Find support, i.e., consistent data points, of the model with parameters $\theta$.} 
\end{algorithmic}
\caption{\bf Outline of Progressive NAPSAC. }
\label{alg:post_processing}
\end{algorithm}

\subsection{Relaxation of the termination criterion}

\begin{figure}[t]
	\centering
	\begin{subfigure}[t]{0.495\columnwidth}
	    \includegraphics[width=1.0\columnwidth]{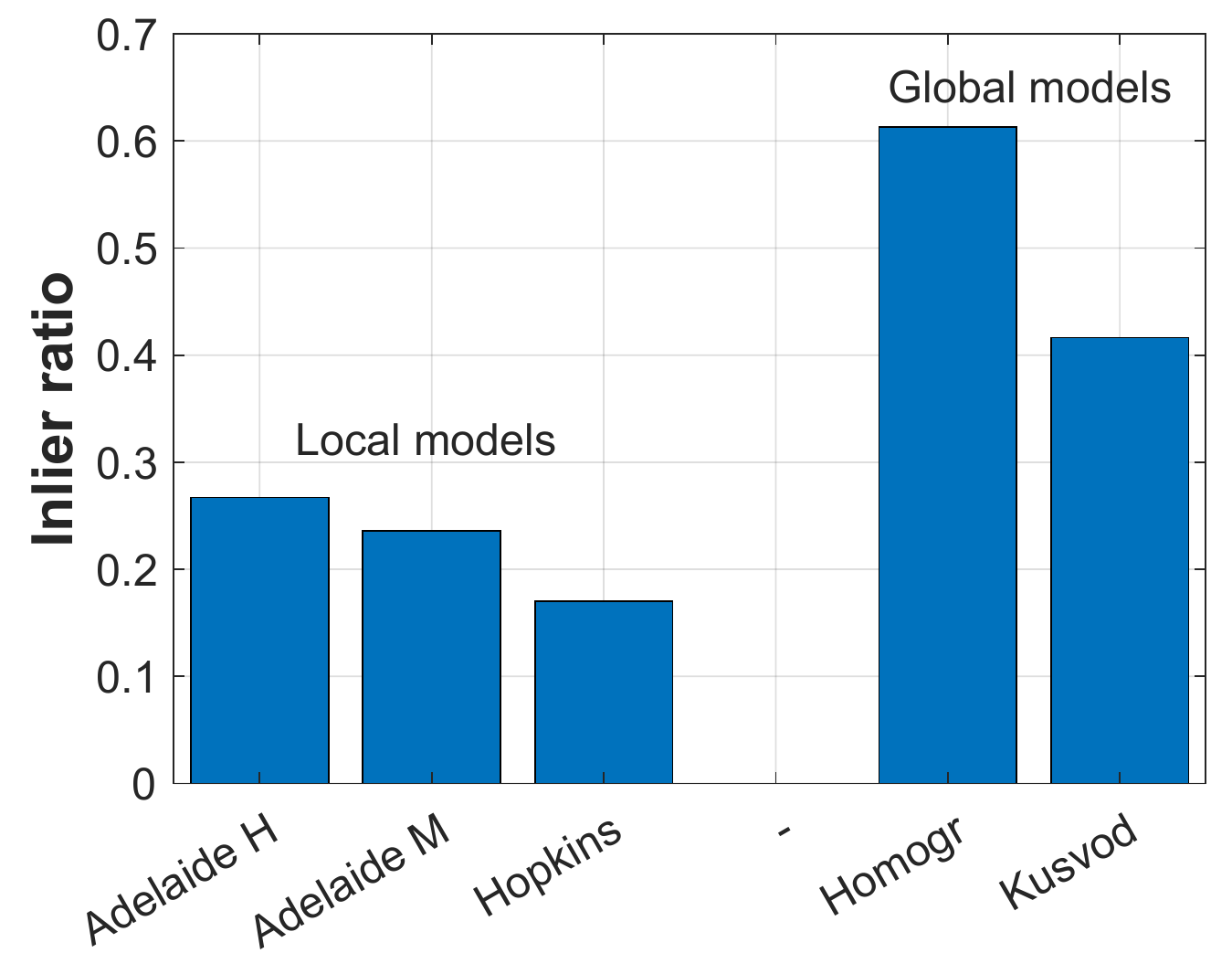}
	    \caption{}
        \label{fig:inlier_ratios}
	\end{subfigure}\hfill
	\begin{subfigure}[t]{0.495\columnwidth}
	    \includegraphics[width=1.0\columnwidth]{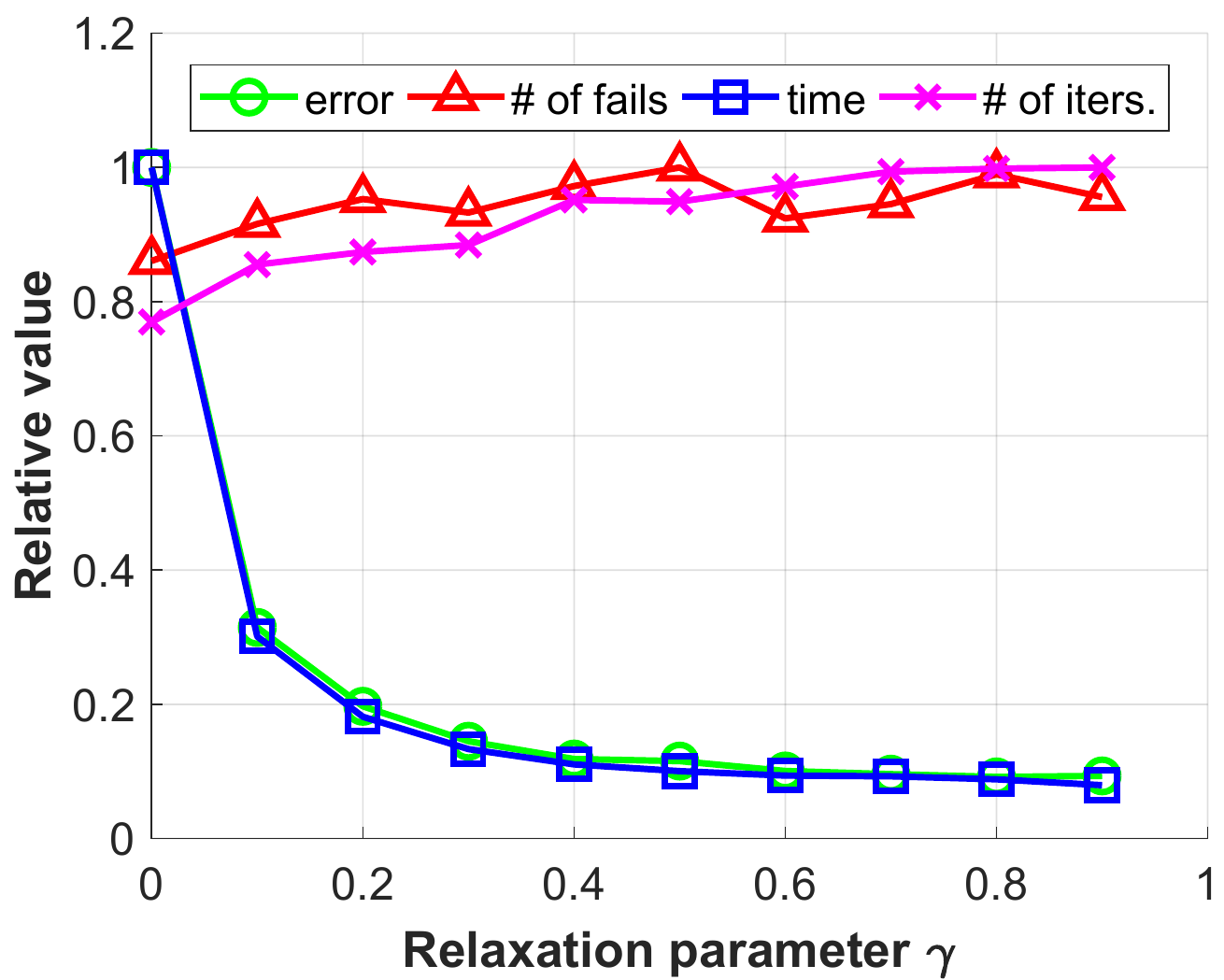}
	    \caption{}
        \label{fig:relaxation}
	\end{subfigure}
    \caption{ (a) The average of the ground truth inlier ratios in five datasets where the models are mostly localized ($1$st--$3$th columns) and where they are not ($4$th--$5$th). 
    (b) The relative (i.e., divided by the maximum) error, number of fails, processing time, and number of iterations are plotted as the function of the relaxation parameter $\gamma$ (from Eq.~\ref{eq:relaxed_ransac_iterations}) of the RANSAC termination criterion. }
\end{figure}

We observed that, in practice, the standard termination criterion proposed for RANSAC is conservative and not suitable for finding local structures early. 
The number of required iterations $t$ of RANSAC is  
\begin{equation}
    t = \frac{\log(1 - \mu)}{\log(1 - \eta^{m})},
    \label{eq:ransac_iterations}
\end{equation}
where $m$ is the size of a minimal sample, $\mu$ is the required confidence in the results and $\eta$ is the inlier ratio.
This criterion does not assume that the points of the sought model are spatially coherent, i.e. that the probablility of selecting a all-inlier sample is higher than $\eta^m$. 
Local structures typically have low inlier ratio as it is demonstrated in Fig.~\ref{fig:inlier_ratios} on different datasets containing local ($1$th--$3$rd columns) and global ($4$th--$5$th) structures. 
Therefore, in the case of low inlier ratio, Eq.~\ref{eq:ransac_iterations} leads to a significant number of iterations even if the model is localized and is found early due to the localized sampling.

There are two simple ways of reducing the number of iterations when having localized models.
The first one is to make assumptions about the data distribution. These distributions nevertheless differ in most of the scenes and, therefore, lead to a more complex problem. 
Another way of terminating earlier is to relax the termination criterion of RANSAC.
It can be easily seen that the number of iterations $t'$ for finding a model with $\eta + \gamma$ inlier ratio is 
\begin{equation}
    t' = \frac{\log(1 - \mu)}{\log(1 - (\eta + \gamma)^{m})},
    \label{eq:relaxed_ransac_iterations}
\end{equation}
where $\gamma \in [0, 1]$ is a relaxation parameter such that $\gamma \leq 1 - \eta$.

\begin{figure*}[h]
    \centering
	\begin{subfigure}[t]{0.50\columnwidth}
	    \includegraphics[width=1.0\columnwidth]{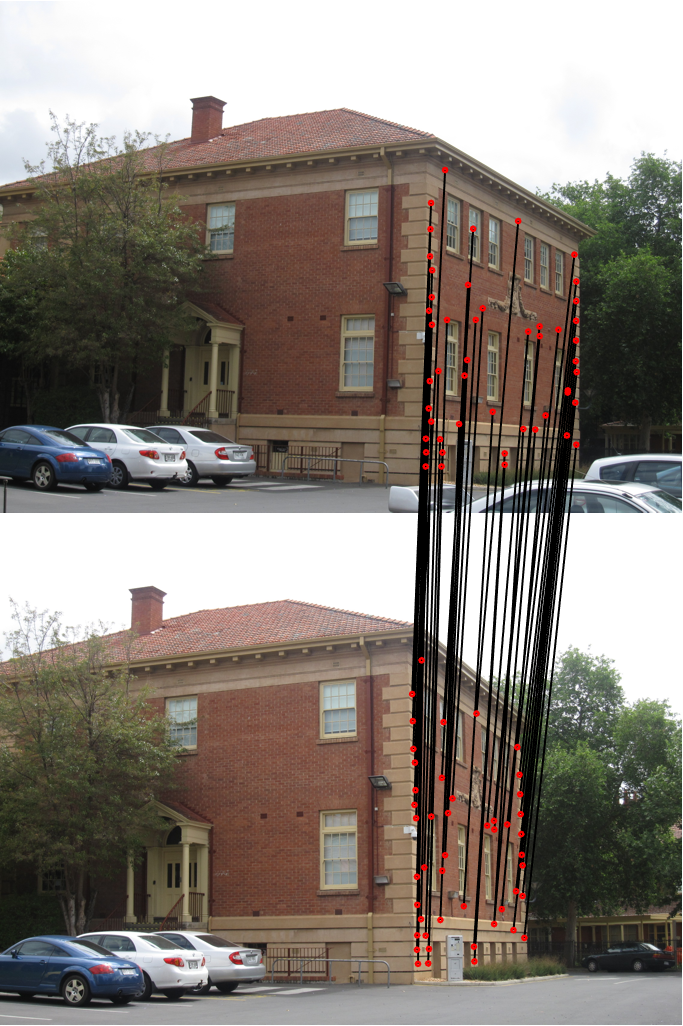}
	    \caption{Ladysymon scene from the {\fontfamily{cmtt}\selectfont AdelaideRMF} {\fontfamily{cmtt}\selectfont homography} dataset.
	    P-NAPSAC made $2\;240$ iterations in $0.05$ secs. PROSAC made $5\;407$ in $0.10$ secs.}
	\end{subfigure}\hfill
	\begin{subfigure}[t]{0.50\columnwidth}
	    \includegraphics[width=1.0\columnwidth]{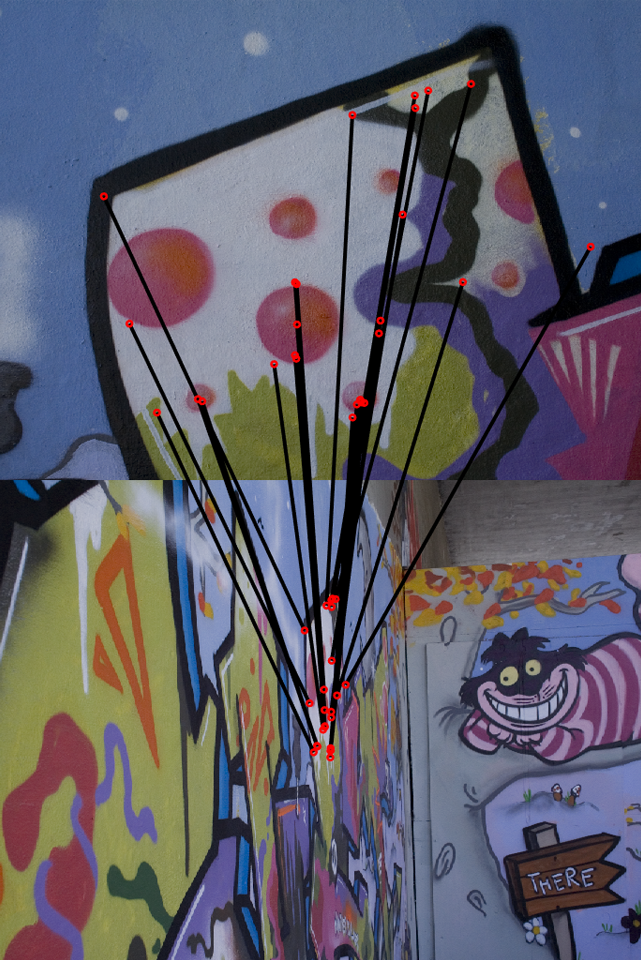}
	    \caption{There scene from the {\fontfamily{cmtt}\selectfont EVD} dataset.
	    P-NAPSAC made $18\;302$ iterations in $0.49$ secs. PROSAC made $84\;831$ in $1.76$ secs.}
	\end{subfigure}\hfill
	\begin{subfigure}[t]{0.50\columnwidth}
	    \includegraphics[width=1.0\columnwidth]{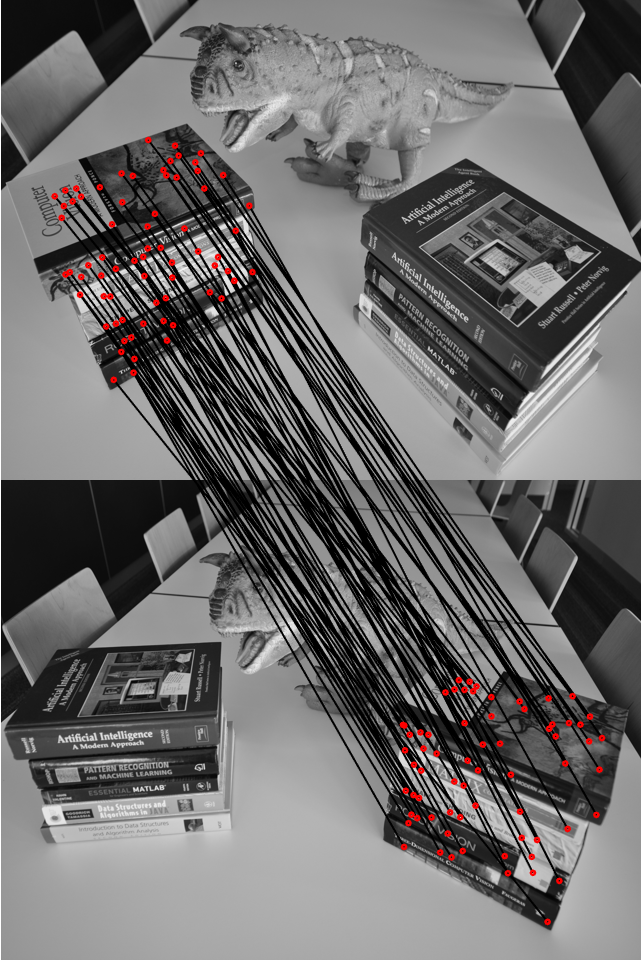}
	    \caption{Dinobooks scene from the {\fontfamily{cmtt}\selectfont AdelaideRMF motion} dataset. 
	    P-NAPSAC made $37\;424$ iterations in $0.65$ secs. PROSAC made $99\;873$ in $1.79$ secs.}
	\end{subfigure}\hfill
	\begin{subfigure}[t]{0.50\columnwidth}
	    \includegraphics[width=1.0\columnwidth]{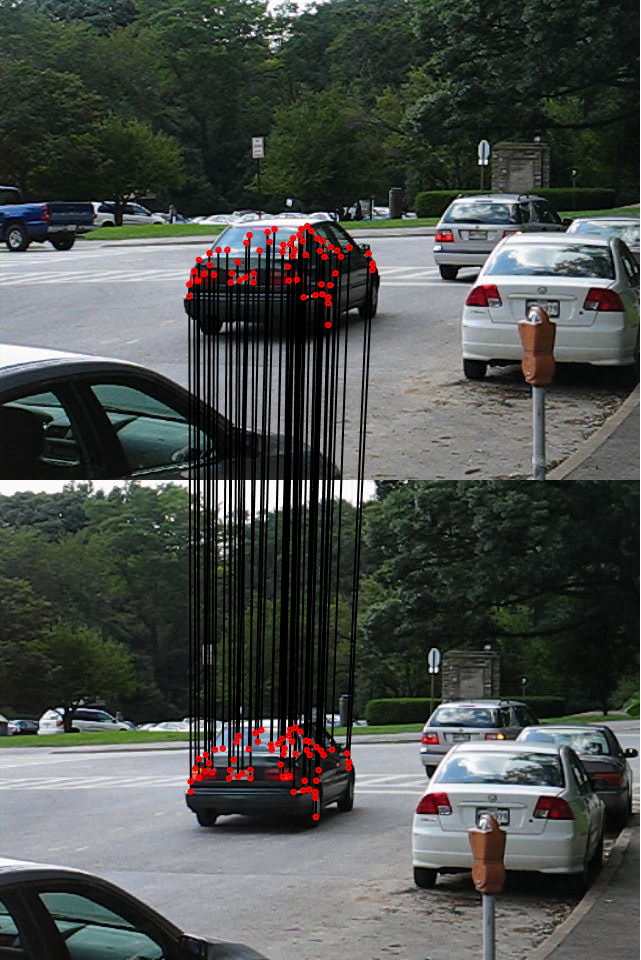}
	    \caption{Cars2 scene from the {\fontfamily{cmtt}\selectfont Hopkins} dataset. 
	    P-NAPSAC made $40197$ iterations in $1.10$ secs. PROSAC made $90672$ in $2.20$ secs. }
	\end{subfigure}\hfill
    \caption{Example image pairs for homography (a--b) and fundamental matrix estimation (c--b) from multiple datasets.
    Inlier correspondences are marked by a line segment joining the corresponding points (red dots).}
    \label{fig:example_result_2}
\end{figure*}

\subsection{Fast neighborhood calculation}

Determining the spatial relations of all points is a time consuming operation even by applying approximating algorithms, e.g., the Fast Approximated Nearest Neighbors method~\cite{muja2009fast}. 
In the sampling of RANSAC-like methods, the primary objective is to find the best sample early and, thus, spending significant time initializing the sampler is not affordable.
Thus, we propose a multi-layer grid for the neighborhood estimation which we describe for point correspondences.
It can be straightforwardly modified considering different input data.

Suppose that we are given two images of size $w_l \times h_l$ ($l \in \{ 1,2 \}$) and a set of point correspondences $\{ (\Point_{i,1}, \Point_{i,2}) \}_{ = 1}^n$, where $\Point_{i,l} = [u_{i,l}, \; v_{i,l}]^\trans$.
A 2D point correspondence can be considered as a point in a four-dimensional space. 
Therefore, the size of a cell in a four-dimensional grid $\mathcal{G}_\delta$ constrained by the sizes of the input image is $\frac{w_1}{\delta} \times \frac{h_1}{\delta} \times \frac{w_2}{\delta} \times \frac{h_2}{\delta}$, where $\delta$ is parameter determining the number of divisions along an axis. 
Function $\Sigma(\mathcal{G}_\delta, [u_{i,1}, \; v_{i,1} \; u_{i,2}, \; v_{i,2}]^\trans)$ returns the set of correspondences which are in the same 4D cell as the $i$-th one. Thus, $|\Sigma(\mathcal{G}_\delta, ...)|$ is the cardinality of the neighborhood of a particular point.
Having multiple layers means that we are given a sequence of $\delta$s such that: $\delta_1 > \delta_2 > ... \ > \delta_d \geq 1$. For each $\delta$, the corresponding $\mathcal{G}_{\delta_k}$ grid is constructed.
For the $i$-th correspondence during its $t_i$-th selection, the finest layer $\mathcal{G}_{\delta_{\max}}$ is selected which has enough points in the cell in which $\Point_i$ is stored. 
Parameter $\delta_{\max}$ is calculated as $\delta_{\max} := \max \{ \delta_k \; : \; k \in [1, d] \wedge |\mathcal{S}_{i, g(t_i) - 1}| \leq |\Sigma(\mathcal{G}_{\delta_k}, ...)| \}$.

In P-NAPSAC, $d = 5$, $\delta_1 = 16$, $\delta_2 = 8$, $\delta_3 = 4$, $\delta_4 = 2$ and $\delta_5 = 1$. When using hash-maps and an appropriate hashing function, the implied computational complexity of the grid creation is $\mathcal{O}(4 n)$. For the search, it is $\mathcal{O}(1)$. Note that $\delta_5 = 1$ leads to a grid with a single cell and, therefore, does not require computation. 

\subsection{\usac}

The Universal Framework for Random Sample Consensus~\cite{raguram2013usac} method combines the state-of-the-art RANSAC components.
In USAC\footnote{\url{http://www.cs.unc.edu/~rraguram/usac/}}, the SPRT test~\cite{chum2008optimal} inspired by Wald's theory is applied for early termination. 
DEGENSAC~\cite{chum2005two} is used for degeneracy testing. 
For optimizing the parameters of the so-far-the-best models, LO-RANSAC~\cite{chum2003locally} is used. 
The model quality is measured by MSAC~\cite{torr2002bayesian}.
PROSAC~\cite{chum2005matching} is applied for sampling. 

In order to update the framework, we replace LO-RANSAC by Graph-Cut RANSAC\footnote{\url{https://github.com/danini/graph-cut-ransac}}~\cite{barath2018graph} which takes the point proximities into account when locally optimizing the parameters of each so-far-the-best model.
Unlike in the original paper~\cite{barath2018graph}, we use the proposed multi-layer grid for determining the point neighborhoods instead of FLANN~\cite{muja2009fast}.
Moreover, we combine P-NAPSAC with PROSAC, such that the first point is selected by PROSAC and the rest of the sample by P-NAPSAC.

\section{Experimental Results}

In this section, the proposed Progressive NAPSAC (P-NAPSAC) sampling is tested on a number of publicly available real-world datasets on homography and fundamental matrix fitting. 
Every evaluated sampler is included into \usac  and was applied using fixed parameters for each problem minimizing the average failure ratio. 
The inlier-outlier threshold was set to 1 pixel for fundamental matrix fitting and 3.2 pixels for homography estimation.
These thresholds were set by exhaustive experimentation, i.e., testing all thresholds in-between $0.5$ and $4.0$ pixels with $0.1$ step size ($0.5$, $0.6$, $...$, $4.0$), applying \usac with every sampler $1\;000$ times. Finally, the threshold was chosen which minimizes the average failure ratio over all tests. 

The evaluated samplers are the uniform sampler of RANSAC~\cite{fischler1981random}, NAPSAC~\cite{nasuto2002napsac} and PROSAC~\cite{chum2005matching}.
Since both the proposed P-NAPSAC and NAPSAC assumes the inliers to be localized, they used the relaxed termination criterion with $\gamma = 0.1$. Thus, they terminate when the probability of finding a model which leads to at least $0.1$ increment in the inlier ratio falls below a threshold. 
Note that this choice will be experimentally justified later. 
PROSAC used its original termination criterion and the quality function for sorting the correspondences was the one proposed in~\cite{chum2005matching}.

Example image pairs for homography (a--b) and fundamental matrix estimation (c--d) from multiple datasets are shown in Fig.~\ref{fig:example_result_2}.
Inlier correspondences are marked by a line segment joining the corresponding points. The numbers of iterations and processing times of PROSAC and P-NAPSAC are reported in the captions of the image pairs. In all cases, P-NAPSAC does significantly fewer iterations than PROSAC. 
Therefore, USAC is speeded up. 

\begin{figure}[h]
    \centering
    \includegraphics[width=0.70\columnwidth]{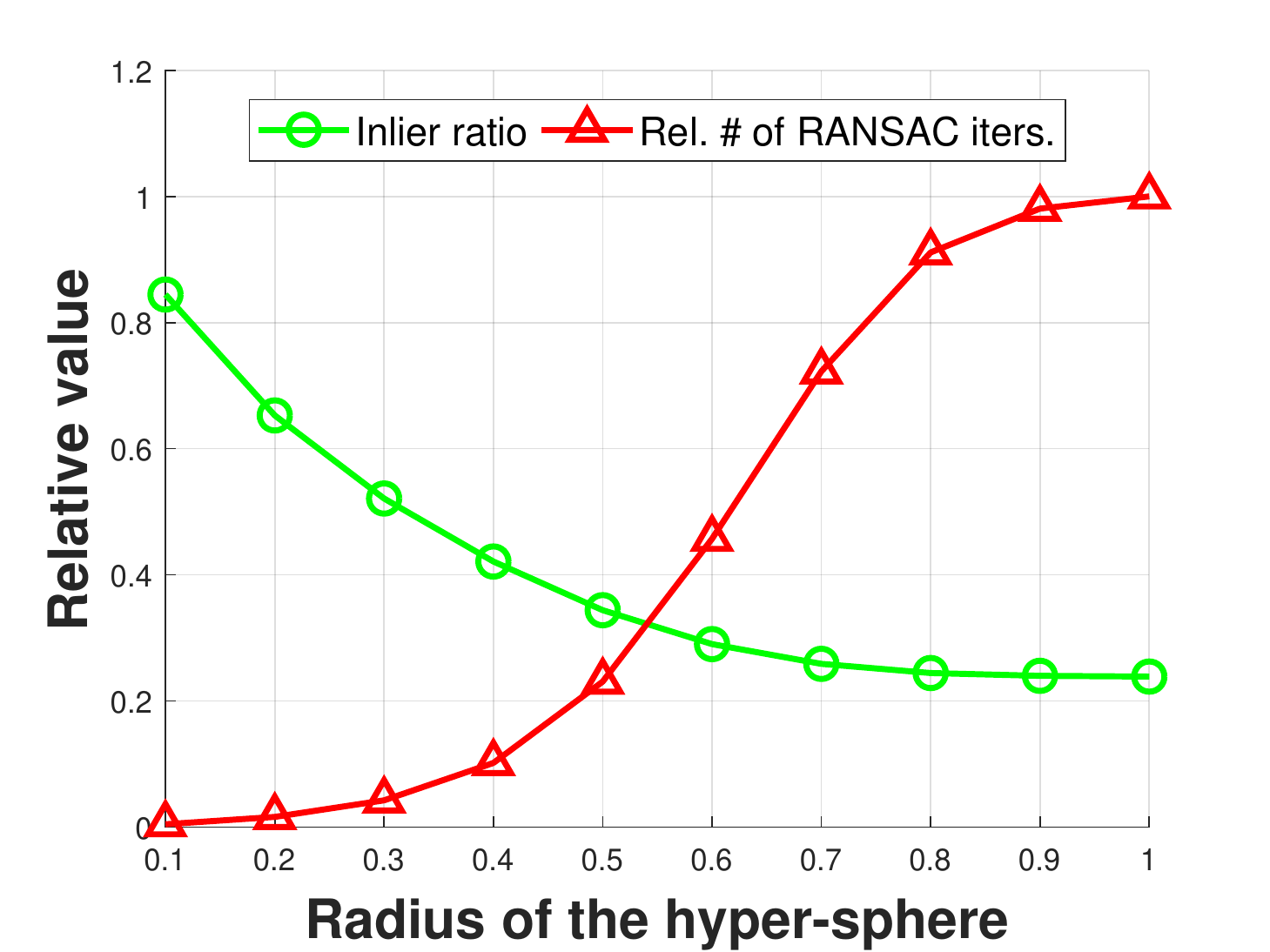}
    \caption{Inlier ratio and rel.\ number of RANSAC iterations, i.e.\ the time w.r.t.\ RANSAC on all points, plotted as the function of radius $r$. The values are averaged over every possible hyper-sphere with an inlier in its center. 
    All image pairs were used from the {\fontfamily{cmtt}\selectfont AdelaideRMF homography} dataset, with $7\;100$ inlier correspondences in total. Radius $r = 1$ means that all points are covered. 
    RANSAC makes 6 iterations when $r = 0.1$ and $1\;416$ when $r = 1.0$.}
    \label{fig:inside_the_ball_H}
\end{figure}

In Fig.~\ref{fig:inside_the_ball_H}, the inlier ratio and the relative number of RANSAC iterations (at $0.99$ confidence) are plotted as the function of the radius $r$ of the hyper-sphere. 
The values are averaged over all possible hyper-spheres with an inlier in the center selected from the provided correspondence sets of each image pair from the {\fontfamily{cmtt}\selectfont AdelaideRMF homography} dataset. In total, $7\;100$ inlier correspondences were used. 
The hyper-sphere is in the correspondence space and, thus, is 4D. 
Radius $r = 1$ means that the ball is big enough to cover the images (i.e., correspondence space) and, therefore, contain all correspondences. 
When $r$ is $0.1$, the theoretical number of RANSAC iterations is $6$. 
When $r$ is $1.0$, the theoretical number of RANSAC iterations is $1\;416$. 
Therefore, the locality assumption provably holds on the {\fontfamily{cmtt}\selectfont AdelaideRMF homography} dataset.

\begin{table*}
	\center
	\resizebox{0.99\textwidth}{!}{
	\begin{tabular}{ r | r | r || c | c | c | c | c | c | c || c }
    \hline
   		\multicolumn{3}{c ||}{\cellcolor{black!20}} & \multicolumn{3}{c | }{\cellcolor{black!20}Homography} & \multicolumn{4}{c ||}{\cellcolor{black!20}Two-view motion (F)} & \cellcolor{black!20} \\
    \hline
   		\multicolumn{3}{c ||}{\cellcolor{black!5}} & \cellcolor{black!5}Adelaide H & \cellcolor{black!5}Homogr & \cellcolor{black!5}EVD & \cellcolor{black!5}Adelaide H & \cellcolor{black!5}Adelaide F & \cellcolor{black!5}Hopkins & \cellcolor{black!5}Kusvod & \cellcolor{black!5}All \\
   	\hline
   		\multicolumn{3}{r ||}{\cellcolor{black!5}\# of models tested} & \cellcolor{black!5}55 & \cellcolor{black!5}16 & \cellcolor{black!5}15 & \cellcolor{black!5}18 & \cellcolor{black!5}40 & \cellcolor{black!5}10\;531 & \cellcolor{black!5}16 & \cellcolor{black!5}10\;691 \\
   	\hline
   		\multicolumn{3}{r ||}{\cellcolor{black!5}local structures?} & \cellcolor{black!5}yes & \cellcolor{black!5}no & \cellcolor{black!5}no & \cellcolor{black!5}no & \cellcolor{black!5}yes & \cellcolor{black!5}yes & \cellcolor{black!5}no & \cellcolor{black!5} \\
    \hline
   		 \multirow{20}{*}{\rotatebox{90}{\usac + }}  & \multirow{5}{*}{\rotatebox{90}{PROSAC}\phantom{x}} & $\epsilon$ (px) & \ph\ph2.6 & \ph1.7 & \ph\textbf{5.5} & \ph\textbf{1.1} & \ph1.0 & \ph\ph1.4 & \ph\ph2.1 & \ph\ph\textbf{2.2}  \\
   		 & & \% of inliers & \ph15.4 & \textbf{31.6} & \ph\textbf{9.6} & 44.5 & 29.0 & \ph40.9 & \ph30.9 & \ph28.8 \\
   		 & & \% of fails & \ph16.6 & \ph\textbf{0.0} & \textbf{19.4} & \ph0.4 & \ph\textbf{0.7} & \ph\ph8.6 & \ph11.8 & \ph\ph8.2 \\
   		 & & $t$ (ms) & 326.1 & \ph\textbf{4.3} & 890.0 & 37.5 & 643.4 & 335.7 & 288.4 & 360.8 \\
   		 & & \# of iters. & 24\;988 & \textbf{172} & 69\;980 & 2\;008 & 43\;049 & 17\;290 & 15\;247 & 24\;676 \\
    \cline{2-11}
   		 & \multirow{5}{*}{\rotatebox{90}{NAPSAC}\phantom{x}} & $\epsilon$ (px) & \ph\textbf{2.0} & \ph\textbf{1.6} & \ph6.4 & \ph2.2 & \ph1.8 & \ph\ph\ph1.7 & \ph\ph3.8 & \ph\ph2.8 \\
   		 & & \% of inliers & 13.9 & 28.6 & \ph6.1 & 30.1 & 25.5 & \ph38.8 & \ph28.6 & \ph24.5 \\
   		  & & \% of fails & 14.7 & \ph\textbf{0.0} & 31.0 & 10.9 & 12.8 & \ph\;\ph\ph9.1 & \ph46.8 & \ph17.9 \\
   		  & & $t$ (ms) & 89.6 & 11.4 & 528.2 & 665.1 & 991.7 & 1\;737.8 & 899.4 & 703.3 \\
   		  & & \# of iters. & 5\;558 & 673 & 37\;603 & 29\;110 & 51\;622 & 27\;589 & 43\;879 & 27\;862 \\
    \cline{2-11}
   		 & \multirow{5}{*}{\rotatebox{90}{\textbf{P-NAPSAC}}\phantom{x}} & $\epsilon$ (px) & \ph2.1 & \ph\textbf{1.6} & \ph5.7 & \ph1.5 & \ph\textbf{0.6} & \ph\ph1.3 & \ph\ph2.3 & \ph\ph\textbf{2.2} \\
   		 & & \% of inliers & \textbf{15.5} & 30.4 & \ph7.5 & 50.3 & 30.8 & \ph42.0 & \ph40.2 & \ph31.0 \\
   		 &  & \% of fails & 15.2 & \ph\textbf{0.0} & 21.1 & \ph\textbf{0.0} & \ph1.0 & \ph\ph7.3 & \ph12.0 & \ph\ph\textbf{8.1} \\
   		 &  & $t$ (ms) & \textbf{74.9} & 10.5 & \textbf{396.2} & \textbf{32.3} & \textbf{602.3} & \textbf{305.9} & \textbf{106.6} & \textbf{218.4} \\
   		 &  & \# of iters. & \textbf{5\;193} &521 & \textbf{30\;889} & \textbf{1\;714} & \textbf{40\;111} & \textbf{16\;479} & \textbf{12\;902} & \textbf{15\;401}\\
    \cline{2-11}
   		 & \multirow{5}{*}{\rotatebox{90}{Uniform}\phantom{x}} & $\epsilon$ (px) & \ph2.7 & \ph\textbf{1.6} & \ph5.8 & \ph\textbf{1.1} & \ph1.1 & \ph\ph\textbf{1.1} & \ph\ph\textbf{1.7} & \ph\ph\textbf{2.2} \\
   		&  & \% of inliers & 15.3 & 25.0 & \ph8.3 & \textbf{58.4} & \textbf{31.8} & \ph\textbf{42.5} & \ph\textbf{41.0} & \ph\textbf{31.8} \\
   		 &  & \% of fails & \textbf{13.1} & \ph\textbf{0.0} & 21.6 & \ph\textbf{0.0} & \ph7.3 & \ph\ph\textbf{6.1} & \ph\textbf{11.3} & \ph\ph8.5 \\
   		 &  & $t$ (ms) & 449.5 & 23.9 & 946.3 & 74.7 & 690.2 & 595.5 & 391.7 & 453.1 \\
   		 &  & \# of iters. & 34\;425 & 1\;716 & 74\;027 & 4\;531 & 45\;886 & 23\;253 & 23\;760 & 29\;657 \\
    \hline
    
    \hline
    \hline      
\end{tabular}}
    \caption{ \textit{Comparison of samplers combined with {\usac}on real-world datasets.} 
    The 1st row shows the problem (i.e., homography or fundamental matrix fitting). 
    The 2nd one reports the datasets (columns) used.
    The numbers of models used in the evaluation are written in the 3rd row for each dataset.
    The 4th one indicates if the datasets contain mostly localized models or not. 
    From the 5th row, each block, consisting of five rows, reports the results of a sampler. 
    The investigated properties are: (i) the average (over $1\;000$ runs on each model) re-projection error in pixels ($\epsilon$) of the estimated homographies w.r.t.\ the inliers provided in the datasets; 
    (ii) the average inlier ratio of the found models (in $\%$), 
    (iii) the frequency of failures ($\%$ of fails); 
    (iv) the processing time ($t$) in milliseconds; 
    (v) and the number of iterations required ($\#$ of iters).
    A run is considered a failure if fewer than the $50\%$ of the ground truth inliers are found.
    Each sampler was combined with USAC which used fixed parameters for each problem minimizing the average failure ratio. 
    The inlier-outlier threshold was set to $1$ pixel for fundamental matrix and $3.2$ pixels for homography estimation.}
    \label{tab:summary_table}
\end{table*}

\vspace{-0.2cm}
\paragraph{Homographies.}  

To test homography estimation, we downloaded {\fontfamily{cmtt}\selectfont homogr} (16 pairs), {\fontfamily{cmtt}\selectfont EVD}\footnote{\url{http://cmp.felk.cvut.cz/wbs/}} (15 pairs) and the {\fontfamily{cmtt}\selectfont AdelaideRMF homography}\footnote{\url{cs.adelaide.edu.au/~hwong/doku.php?id=data}} (19 pairs) datasets. Each consists of image pairs of different sizes from $329 \times 278$ up to $1712 \times 1712$ with point correspondences and inliers selected manually.   

The {\fontfamily{cmtt}\selectfont Homogr} dataset consists of mostly short baseline stereo images, whilst the pairs of {\fontfamily{cmtt}\selectfont EVD} undergo an extreme view change, i.e., wide baseline or extreme zoom. In both datasets, the correspondences are assigned manually to one of the two classes, i.e., outlier or inlier of the most dominant homography in the scene. 
In the {\fontfamily{cmtt}\selectfont Homogr} dataset the models are not localized.
The {\fontfamily{cmtt}\selectfont EVD} dataset contains a mixture of local and global models.  

In the image pairs of the {\fontfamily{cmtt}\selectfont AdelaideRMF homography} dataset, the provided correspondences are assigned to multiple homographies or the outlier class. 
Given an image pair, the procedure to evaluate the samplers is the following.
First, the ground truth homographies, estimated from the manually annotated correspondences, are selected one after another. 
For each homography in the annotated set, the procedure is as follows: 
\begin{enumerate}
    \item The correspondences which do not belong to the selected homography are replaced by completely random correspondences (inside the image boundaries) to reduce the probability of finding a different model than what is currently tested.  
    \item \usac is combined with each competitor sampler and is applied to the point set consisting of the inliers of the current homography and outliers.
    \item The estimated homography is compared to the manually selected inliers provided in the datasets.
\end{enumerate}
All algorithms applied the normalized four-point algorithm~\cite{hartley2003multiple} for homography estimation both in the model generation and local optimization steps. 
In these images, the points originate mostly from the walls of buildings and, thus, form spatially coherent structures. 

In Table~\ref{tab:summary_table}, the first row shows the problem (i.e., homography or fundamental matrix fitting). 
The second one reports the datasets (columns) used.
The numbers of models, on which the methods were tested, are written in the third row for each dataset.
The fourth one indicates if the datasets contain mostly localized models or not. 
From the fifth row, each block, consisting of five rows, reports the results of a sampler. 
The investigated properties are: (i) the average re-projection error in pixels ($\epsilon$) of the estimated homographies with respect to the inliers provided in the datasets; 
(ii) the average inlier ratio, in percentage, of the found models w.r.t.\ the entire point set; 
(iii) the frequency of failures (percentage of fails); 
(iv) the processing time ($t$) in milliseconds; 
(v) and the number of iterations required ($\#$ of iters).
A run is considered a failure if fewer than the $50\%$ of the ground truth inliers are found.
The values are averaged over $1\;000$ runs on each tested model. 

It can be seen that the re-projection error, inlier ratio and failure rate of P-NAPSAC are similar to that of PROSAC.
Thus the estimated models are of the same quality. 
However, P-NAPSAC \textit{requires the fewest iterations on two out of the three datasets} and, therefore, its processing time is the lowest (by 2-3 times compared to PROSAC) on them.
On the {\fontfamily{cmtt}\selectfont Homogr} dataset, it is the second fastest behind PROSAC by merely $6.2$ milliseconds. 

\vspace{-0.2cm}
\paragraph{Fundamental Matrices.}  

To evaluate the performance on fundamental matrix estimation we downloaded {\fontfamily{cmtt}\selectfont kusvod2}\footnote{\url{http://cmp.felk.cvut.cz/data/geometry2view/}} (24 pairs),  {\fontfamily{cmtt}\selectfont AdelaideRMF homography} (19 pairs),  {\fontfamily{cmtt}\selectfont AdelaideRMF motion} (19 pairs), and {\fontfamily{cmtt}\selectfont hopkins}\footnote{\url{http://www.vision.jhu.edu/data/hopkins155/}} datasets. 

{\fontfamily{cmtt}\selectfont Kusvod2} consists of 24 image pairs of different sizes with point correspondences assigned to the dominant rigid motion, i.e., fundamental matrix, or to the outlier class manually.
For {\fontfamily{cmtt}\selectfont AdelaideRMF homography} dataset, all of the points assigned to a homography are considered as the inliers of the rigid motion regarding to the background. 
In these datasets, the model corresponds to the background motion and, therefore, its correspondences are not localized. 

The {\fontfamily{cmtt}\selectfont AdelaideRMF motion} dataset consists a total of 19 image pairs with point correspondences, each assigned manually to a rigid motion or the outlier class.
The {\fontfamily{cmtt}\selectfont hopkins} dataset consists of $155$ video sequences with point trajectories provided, each assigned to a rigid motion.  
For both datasets, we applied the procedure explained in the previous section to select and test the models one by one. For the {\fontfamily{cmtt}\selectfont hopkins} dataset, this procedure was done for every possible image pairs in each video sequence. The models are spatially coherent. 
The procedure resulted in a total of $10\;571$ tested models.

All methods applied the seven-point method~\cite{hartley2003multiple} as a minimal solver for estimating the fundamental matrix. Thus they drew minimal samples of size seven in each iteration. All fundamental matrices were discarded for which the oriented epipolar constraint~\cite{chum2004epipolar} did not hold. For the final least squares fitting, the normalized eight-point algorithm~\cite{hartley1997defense} was run on the obtained inlier set.

In Table~\ref{tab:summary_table}, the columns from the $6$-th to $9$-th report the results of fundamental matrix fitting. 
It can be seen that the re-projection error, inlier ratio and failure rate of P-NAPSAC are similar to that of the other methods -- sometimes better, sometimes slightly worse.
\textit{P-NAPSAC lead to the earliest termination on all evaluated datasets.}

The last column of Table~\ref{tab:summary_table} summarizes the results on both investigated problems. 
It can be seen that P-NAPSAC, PROSAC and Uniform sampling lead to the same geometric error.
In terms of inlier ratio, uniform sampling is slightly ahead of the second best P-NAPSAC (by 0.8\%). However, P-NAPSAC requires half the processing time of uniform sampling.
The failure ratio of P-NAPSAC is the lowest by a margin of $0.1$ percentage.
\textit{On average, P-NAPSAC leads to the fast robust estimation.}
It makes $0.66$ times fewer iterations than PROSAC which is the second fastest sampler amongst the compared methods.
Therefore, \textit{USAC is significantly speeded up by using P-NAPSAC as the sampler.} 

\vspace{-0.2cm}
\paragraph{Relaxed termination criterion.}

In order to test the relaxed termination criterion, we applied Progressive NAPSAC to all datasets with different $\gamma$s. 
We then investigated how each property (i.e., the error of the estimated model, failure rate, processing time, and number of iterations) changes.
Fig.~\ref{fig:relaxation} plots the average of the reported properties as the function of $\gamma$ (over $100$ runs on each scene). 
The relative values are shown. Thus, for each test, the values are divided by the maximum. 
For instance, if P-NAPSAC draws $100$ iterations when $\lambda = 0$, the number of iterations is divided by $100$ for every other $\lambda$. 

It can be seen that the error and failure ratio slowly increase from approx.\ $0.8$ to $1.0$. The trend seems to be close to linear. 
Simultaneously, the number of iterations and, thus, the processing time are reduced.
It can, however, be observed that the trend is not linear. 
Around $\lambda = 0.1$ there is significant drop from $1.0$ to $0.3$. 
If $\lambda > 0.1$ both values decrease mildly.
Therefore, selecting $\lambda = 0.1$ as the relaxation factor does not lead to noticeably worse results but speeds up the procedure significantly.

\section{Conclusion}

In this paper, we propose Progressive NAPSAC combining the two strands of sampling (local and global) used in RANSAC-like robust estimators. 
Considering that nearby points often originate from the same model in real-world data, P-NAPSAC finds local structures earlier than global samplers.
In addition, by blending progressively from local to global sampling, it does not suffer from the weaknesses of purely localized samplers. In P-NAPSAC, the blending factor is a function of the data. 
Moreover, USAC is updated by GC-RANSAC and the proposed sampler.
P-NAPSAC is tested on homography and fundamental matrix fitting on a total of $10\;691$ models from 7 publicly available datasets and is reported to trigger earlier termination of \usac than state-of-the-art samplers. 

{\small
\bibliographystyle{ieee}
\bibliography{egbib}
}

\end{document}